\documentclass[10pt]{article}
\usepackage{multicol,setspace}
\usepackage{hyperref}
\usepackage{fontawesome}
\usepackage{amsmath}
\usepackage{amssymb}
\usepackage{nccmath}
\usepackage{authblk}
\usepackage{fancyhdr}
\usepackage{picins}
\usepackage{lastpage}
\usepackage{geometry}
\usepackage{orcidlink}
\usepackage{caption}
\usepackage{enumitem}
\usepackage{graphicx}
\usepackage{nomencl}
\usepackage[numbers,longnamesfirst]{natbib}
\usepackage[document]{ragged2e}
\usepackage[font=small,labelfont=bf]{caption}
\usepackage{bm}
\newcommand{\mat}[1]{\bm{\mathrm{#1}}}
\usepackage{todonotes}
\definecolor{darkgreen}{rgb}{0, .5, 0}

\hypersetup{
  colorlinks = true,
  linkcolor=blue,
  linkbordercolor= {0 0 1}}
  
\providecommand{\keywords}[1]
{
  \textbf{Keywords:} #1
}

\makeatletter
\newenvironment{tablehere}
  {\def\@captype{table}}
  {}

\newenvironment{figurehere}
  {\def\@captype{figure}}
  {}
\makeatother

\begin{document}

\justifying
\makenomenclature

\title{\vspace{-2.5em} \normalsize IAC-23,D1,6,1,x78669 \vspace{1em} \\ \normalsize \textbf{Multi-Agent 3D Map Reconstruction and Change Detection in Microgravity with Free-Flying Robots}}

\author[1\textsection]{\textbf{\normalsize{Holly Dinkel}}\orcidlink{0000-0002-7510-2066}}
\author[2\textsection]{\textbf{\normalsize{Julia Di}}\orcidlink{0000-0001-5872-5694}}
\author[3]{\textbf{\normalsize{Jamie Santos}}\orcidlink{0009-0007-7923-8482}}
\author[4]{\textbf{\normalsize{Keenan Albee}}\orcidlink{0000-0002-9655-2429}}
\author[5]{\textbf{\normalsize{Paulo Borges}}\orcidlink{0000-0001-8137-7245}}
\author[6]{\\\textbf{\normalsize{Marina Moreira}}\orcidlink{0000-0002-4180-473X}}
\author[6]{\textbf{\normalsize{Oleg Alexandrov}}\orcidlink{0000-0001-7567-493X}}
\author[6]{\textbf{\normalsize{Brian Coltin}}\orcidlink{0000-0003-2228-6815}}
\author[6]{\textbf{\normalsize{Trey Smith}}\orcidlink{0000-0001-8650-8566}}

\affil[1]{\normalsize{\textit{Department of Aerospace Engineering at the University of Illinois Urbana-Champaign, Urbana, IL, USA}}}
\affil[2]{\normalsize{\textit{Department of Mechanical Engineering at Stanford University, Stanford, CA, USA}}}
\affil[3]{\normalsize{\textit{Department of Physics at Chalmers University of Technology, Gothenburg, Sweden}}}
\affil[4]{\normalsize{\textit{Maritime and Multi-Agent Autonomy Group at the Jet Propulsion Laboratory, California Institute of Technology, Pasadena, CA, USA}}}
\affil[5]{\normalsize{\textit{Robotics and Autonomous Systems Group at CSIRO Data61, Brisbane, QLD, Australia}}}
\affil[6]{\normalsize{\textit{Intelligent Robotics Group at the NASA Ames Research Center, Moffett Field, CA, USA}}}
\date{\vspace{-5ex}}
\maketitle

\begingroup\renewcommand\thefootnote{\textsection}
\footnotetext{Equal contribution}
\endgroup

\vspace{-2em}

\begin{abstract}
\normalsize 
Assistive free-flyer robots autonomously caring for future crewed outposts---such as NASA’s Astrobee robots on the International Space Station (ISS)---must be able to detect day-to-day interior changes to track inventory, detect and diagnose faults, and monitor the outpost status. This work presents a framework for multi-agent cooperative mapping and change detection to enable robotic maintenance of space outposts. One agent is used to reconstruct a 3D model of the environment from sequences of images and corresponding depth information. Another agent is used to periodically scan the environment for inconsistencies against the 3D model. Change detection is validated after completing the surveys using real image and pose data collected by Astrobee robots in a ground testing environment and from microgravity aboard the ISS. This work outlines the objectives, requirements, and algorithmic modules for the multi-agent reconstruction system, including recommendations for its use by assistive free-flyers aboard future microgravity outposts.
\end{abstract}
\keywords{Change Detection, Microgravity, Robotics, 3D Reconstruction, Mapping, Artemis}

\begin{multicols}{2}
\normalsize
\subsection{Introduction}\vspace{-0.5em}

\begin{figure*}
\centering
\includegraphics[width=0.75\textwidth]{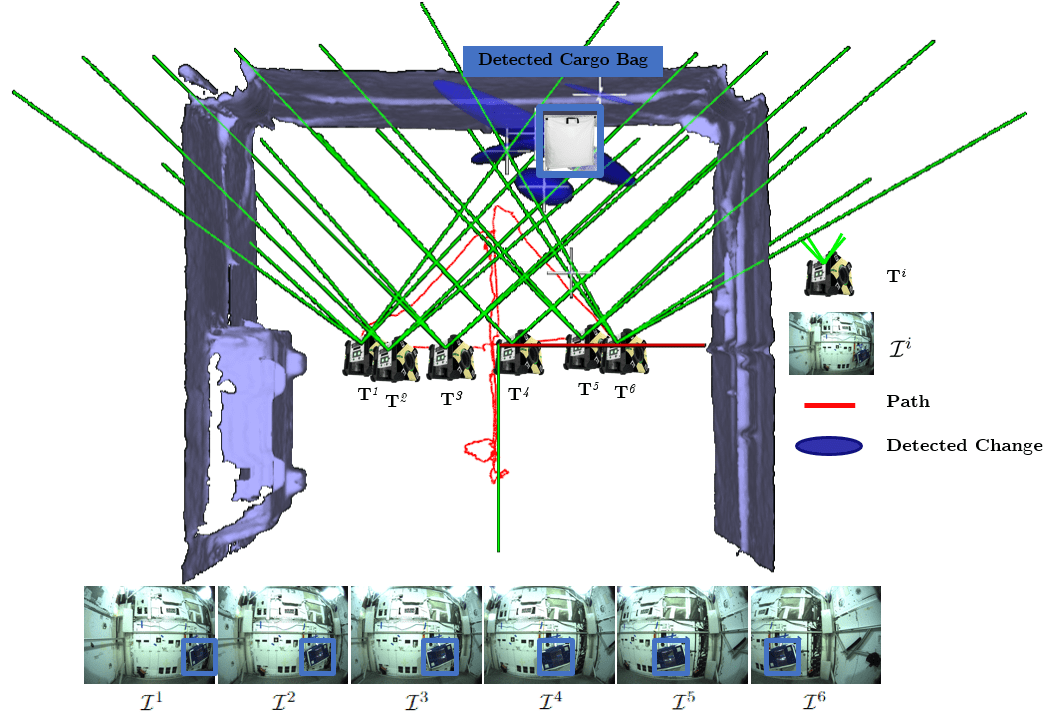}
\caption{\justifying FastCD detects scene changes against a pre-built 3D model of the environment by computing inconsistencies among a set of RGB images projected onto the 3D model using their corresponding poses, $\mat{T}^i$.}
\label{fig: fastcd}
\end{figure*}

On the International Space Station (ISS), astronaut time is expensive and limited. Future orbital outposts, including Gateway, will likely have limited crews or even be uncrewed for extended periods of time. Routine tasks on these outposts provide compelling opportunities for robotic assistance \cite{crusanDeepSpaceGateway2018} \cite{lehnhardt2022gateway}. A number of free-flying assistants have been proposed to address this need including the SPHERES \cite{otero2002spheres}, CIMON, Int-Ball \cite{bualatAstrobeeNewTool2018}, and most recently Astrobee \cite{smith2016astrobee} robots. Astrobee is a next-generation free-flying robot onboard the ISS capable of deploying a variety of future research payloads and guest science software intended to serve as an astronaut assistant to alleviate valuable crew time~\cite{smith2016astrobee,carlino2019astrobee,bualat2021astrobee,astrobee}. One area where Astrobee has the potential to reduce astronaut workload is as a mobile sensor conducting and recording surveys. Astronauts currently conduct manual regular visual surveys of the entire space station, which Astrobee could alleviate. Free-flying robots, including Astrobee, will be increasingly useful assistants, managing and maintaining future microgravity outposts.

However, one challenge for deploying robots in human habitats is the interiors of these environments are ever-changing: objects frequently disappear or are reintroduced. A current map of the environment is vital for maintenance and navigation, but creating a new map and computing between-map changes is resource-intensive as robots could take hours to complete an environment scan. Rather than repeat the mapping process for the entire environment, it is advantageous to only recompute local changes---Fast Image-Based Geometric Change Detection (FastCD) enables robots to quickly locate changes in a known world model based on geometric projection operations performed on a small ($<10$) batch of images as shown in Figure \ref{fig: fastcd} \cite{palazzolo2017change}. This work develops FastCD for operation on assistive free-flyers in microgravity. This work contributes:

\begin{enumerate}
    \item Discussion of a multi-agent scene change detection framework for free-flyer robots. This application can also be extended for object discovery.
    \item Discussion of scene change detection considerations for resource-constrained mobile robots in a space environment.
    \item Demonstration and evaluation of FastCD on data collected with Astrobee robots at the NASA Ames Research Center Granite Lab microgravity research facility and aboard the ISS.
    \item Release of a FastCD-compatible dataset of Granite Lab and ISS activity data at \href{https://bit.ly/astrobee_fastcd_data}{https://bit.ly/astrobee\_fastcd\_data}.
\end{enumerate}

\subsection{Related Work}
Because environment models should be up-to-date for most applications, a common task in robotic surveying is environment reconstruction. For larger environments, obtaining a new scan and generating 3D models of the environment can be expensive, requiring dedicated robot mapping time, sensors, and significant data processing. This cost can be reduced through 3D scene change detection \cite{qin20163d}. By identifying only local changes in an environment based on existing information, which can be generated directly from raw imagery, a mobile robot may direct its surveying to target change regions.

There are several types of change detection methods categorized according to the target scene. Classical change detection is commonly formulated as a 2D or 2.5D problem, comparing new image data to old background images, and is adopted in several satellite remote sensing and surveillance use cases \cite{alcantarilla2018street, celik2009unsupervised, radke2005image}. Pairwise image comparison is straightforward, but may not scale proportionally to map size. For traversing through a 3D environment in robotic outpost maintenance, a 3D change detection approach is better suited.

Within 3D change detection, several works have investigated probabilistic modeling. One recent unsupervised method of scene change detection uses Gaussian Mixture Model (GMM) Clustering to identify changes between complete point cloud maps of the environment~\cite{santos2023unsupervised}. This work processes a complete point cloud map from one survey into region-based GMM clusters through split-and-merge expectation-maximization \cite{li2009novelGMM}, and detects between-scene changes by comparing the cluster sets with the Earth Mover's Distance \cite{rubner2000earth}. Since this method requires complete point cloud maps of the environment, it cannot be deployed online on a mobile platform and is better suited for situational awareness applications rather than anomaly response.

Another recent method repurposed map-based localization for change detection \cite{kanji2019localization}. This method does not perform well if self-localization methods are poor, which may be the case in a space environment. For example, free-flying assistive robots in microgravity do not have the availability of a gravity vector for localization. Furthermore, in dense environments like the ISS, map landmarks are not always available when there are occlusions from cargo bags or human activity. Maps are also expensive to build online for a mobile robot with limited computing power.

Other hybrid methods augment a 2D method with 3D information for remote sensing applications. Changes are detected by comparing unordered 2D images collected by a mobile agent to a 3D building information model which contains information about the geometry of the environment ~\cite{golparvarfard2011monitoring}. This method may work well for construction applications where environments are developed according to a blueprint, but is not well suited for change detection in human habitats which are frequently significantly reconfigured and for which a blueprint may be unavailable.

Finally, other geometric treatments of 3D change detection create inconsistency maps from RGB images or depth projections when compared to 3D models \cite{palazzolo2017change, taneja2011image, taneja2013city}. These methods are fast and lightweight, suitable for operation on a mobile robot when given a 3D world model of the environment, but are sensitive to localization accuracy. 

\subsection{Methodology}
\label{sec:methods}

This work applies FastCD to sequences of RGB images to detect scene changes and discover novel objects. The FastCD system is illustrated in Figure \ref{fig: pipeline} and discussed in the following sections. First, raw image and pose data are collected with the Astrobee robot (Section \ref{sec:data-collection}) and processed to remove low-movement images (Section \ref{sec:low-movement}). A camera pose is estimated for each image (Section \ref{sec:est-cam-pose}). A baseline 3D model of the environment is created (Section \ref{sec:baseline-map}). Finally, candidate inconsistencies are computed between the images used for change detection, and the resulting 3D regions of change are output (Section \ref{sec:inconsistencies}).

\columnbreak

\begin{figurehere}
\footnotesize
\centering
\captionsetup{type=figure}
\includegraphics[width=\columnwidth]{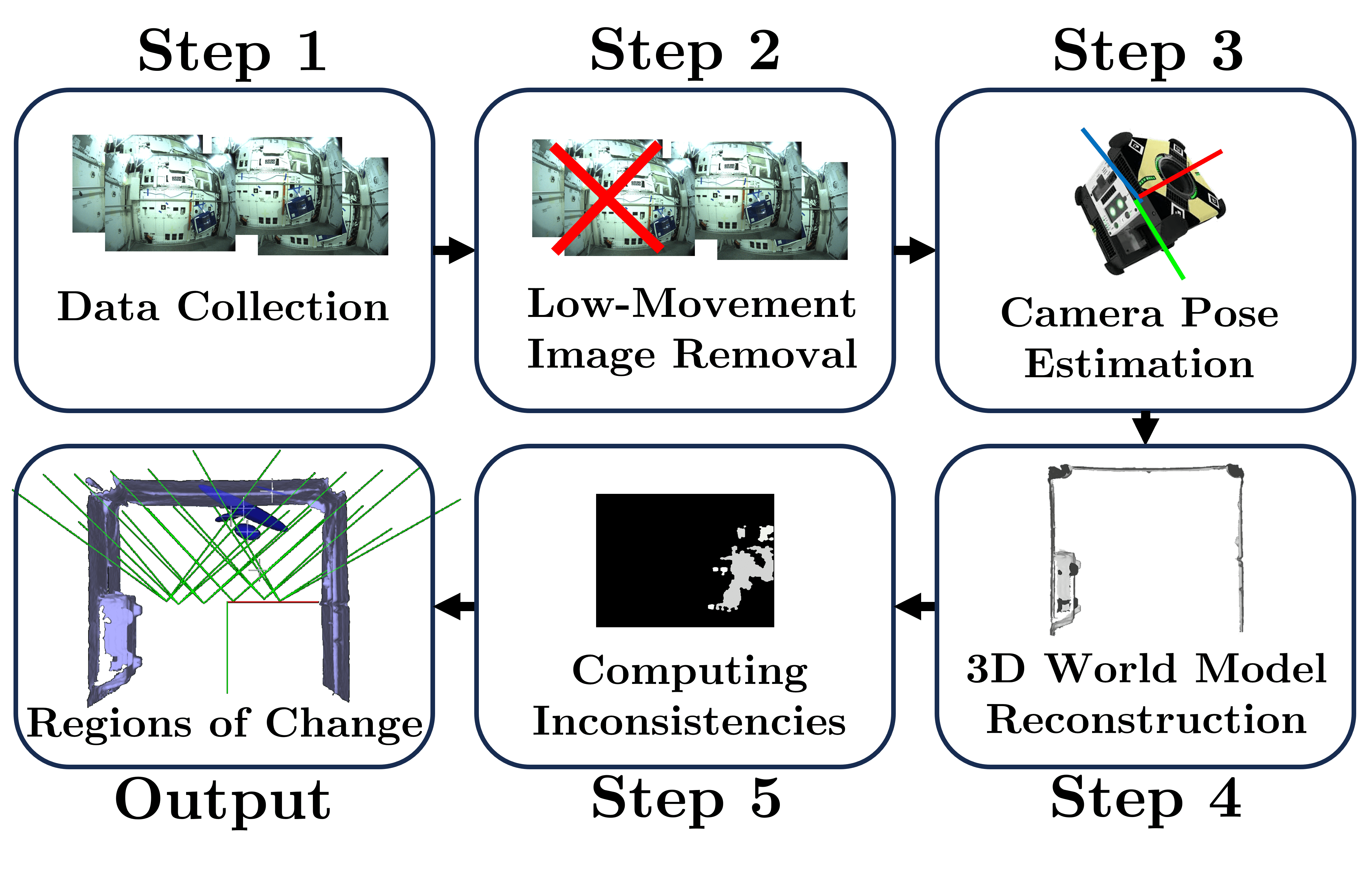}
\caption{\justifying From an image sequence, the change detection system removes low-movement images, estimates the camera pose for remaining images, builds a 3D environment map, computes between-image inconsistencies, and outputs detected changes.}
\label{fig: pipeline}
\vspace{1em}
\end{figurehere}

\begin{figurehere}
\centering
\footnotesize
\captionsetup{type=figure}
\includegraphics[width=0.9\columnwidth]{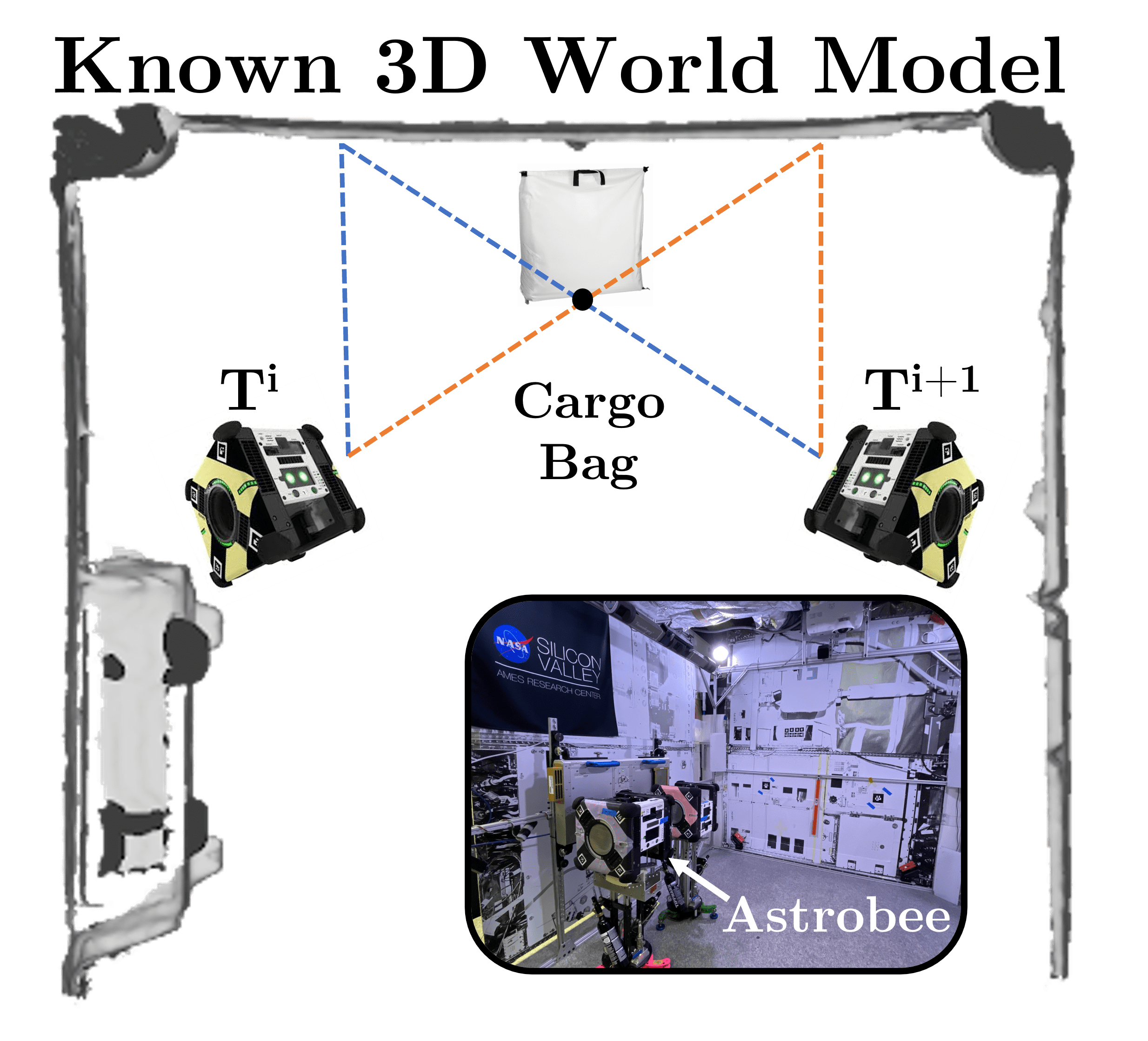}
\caption{\justifying FastCD first back-projects images captured from different poses, $\mat{T}^{i}$ and $\mat{T}^{i+1}$, onto a 3D model. Next, it re-projects the 3D points from the model onto a common image plane for comparison. (Inset) The Granite Lab is an environment which mimics microgravity for testing the ground Astrobee units.}
\label{fig: geometry}
\vspace{1em}
\end{figurehere}

\noindent Figure \ref{fig: geometry} provides geometric intuition for FastCD, summarized in the following steps: 
\begin{enumerate}
    \item Let $i$ represent the index of an image-pose pair in the data sequence.
    \item Back-project (pixel-to-point) image $\mathcal{I}^i$ from pose $\mat{T}^i$ into the World frame of the 3D model using the pose and projection matrix of the camera. 
    \item Re-project (point-to-pixel) image $\mathcal{I}^i$ onto the image plane $\mathcal{I}^{i \rightarrow i+1}$ of the camera at pose $\mat{T}^{i+1}$ to create a new image containing the content from the image captured at $\mat{T}^i$ at pose $\mat{T}^{i+1}$. 
\end{enumerate}

\noindent If there is no change in the 3D model between when the image was acquired and when the model was created, all pixels taken from pose $\mat{T}^{i}$ should correctly re-project onto $\mathcal{I}^{i\rightarrow i+1}$. In other words, the image $\mathcal{I}^{i+1}$ at $\mat{T}^{i+1}$ and the image $\mathcal{I}^{i}$ from $\mat{T}^{i}$ re-projected onto $\mathcal{I}^{i\rightarrow i+1}$ should be approximately identical. If there is a change from what is conveyed in the model, pixels corresponding to the change will re-project onto $\mathcal{I}^{i\rightarrow i+1}$ incorrectly.

\subsubsection{Data Collection}
\label{sec:data-collection}

Data in this work were collected with Astrobee units in the Granite Lab and on the ISS. The Granite Lab (inset in Figure \ref{fig: geometry}) is a facility that replicates visual features of the ISS and mimics 3 DOF microgravity by placing Astrobee on a near-frictionless air bearing.

The Navigation Camera (NavCam) on Astrobee is a fixed-focus RGB camera with a wide field of view. This sensor collects Bayer images (at $1280 \times 960$ resolution) at 5 Hz \cite{smith2016astrobee,astrobee}. When Astrobee surveys its environment, it flies to waypoint stations where it may hold its pose to capture clear, high-resolution ($5344 \times 4008$) images with its Scientific Camera (SciCam). Images collected with the SciCam cannot be streamed due to their large sizes, so this work uses NavCam imagery by default. 

Two types of paths were designed for Astrobee. In the surveys of the Granite Lab, the paths were designed to emphasize translation. In the surveys of the ISS, the paths were designed to emphasize 360$^o$ rotation, approximately fixing the position of Astrobee. Figure \ref{fig: paths} plots two paths against the corresponding model, one for a survey in the Granite Lab and one for a survey in the ISS. Throughout this work, different Astrobee units are referenced by their nicknames (``BSharp'', ``Queen'', and ``Bumble''), as used in the Granite Lab and ISS deployments. 

\subsubsection{Low-Movement Image Removal}
\label{sec:low-movement}

Because Astrobee's flight speed is low compared to the NavCam sensor rate, the robot may capture sequential NavCam images which exhibit little to no movement between-frame. Data are processed by removing these

\columnbreak

\begin{figurehere}
\footnotesize
\centering
\captionsetup{type=figure}
\includegraphics[width=\columnwidth]{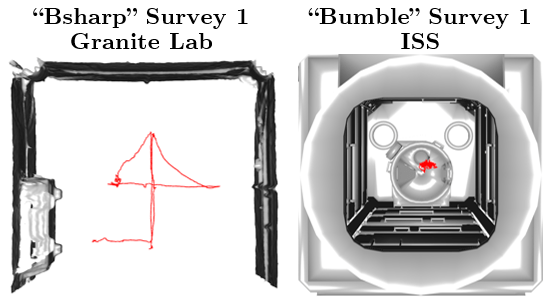}
\caption{\justifying The paths of the Granite Lab surveys emphasize translational wall-scanning and the paths of the ISS surveys emphasize rotation.}
\label{fig: paths}
\vspace{1em}
\end{figurehere}

\noindent low-movement images through feature matching and optical flow. Sparse features are detected and matched between two images using the pyramidal implementation of the Lucas-Kanade optical flow feature tracker \cite{bouguet2001pyramidal}. If the distance between these sets of features is less than a maximum distance threshold, low movement is detected between the images and one of the images is discarded. Low-movement image removal reduces the number of image candidates for change detection from thousands to tens \cite{astrobee}.

\subsubsection{Camera Pose Estimation}
\label{sec:est-cam-pose}

FastCD requires an estimate of the pose of the camera in the World coordinate frame defined by the 3D model. The camera pose is estimated as a homogeneous transformation matrix, $\mat{T}^i_{4 \times 4}$, where $i$ represents the index of the image-pose pair in the data sequence. The pose $\mat{T}^i$ can be computed from the body pose of the Astrobee robot in the World frame, $\mat{T}^i_{\text{Astrobee}}$, and the pose of the NavCam with respect to the Astrobee robot, $^{\text{Astrobee}}\mat{T}_{\text{NavCam}}$. The pose of the Astrobee robot is estimated at a rate of approximately 5 Hz with a factor graph-based localization system using an inertial measurement unit on the robot \cite{dellaert2012factor,carlone2014eliminating,soussan2022astroloc}. The extrinsic calibration of the frame of the NavCam with respect to the body frame of the Astrobee robot was performed using Kalibr \cite{kalibr,furgale2012kalibr,furgale2013kalibr,maye2013self}. The pose is computed as

\begin{align}
    \mat{T}^i = ^{\text{World}}\mat{T}_{\text{Astrobee}}^i & ^{\text{Astrobee}}\mat{T}_{\text{NavCam}}.
\end{align}

\subsubsection{3D World Model Reconstruction}
\label{sec:baseline-map}
The NASA Ames Stereo Pipeline package generates a 3D map of the environment from Astrobee Navigation Camera (NavCam), Hazard Camera (HazCam), and localization data \cite{smith2021isaac,nasa-isaac}. It registers image data from the NavCam with depth information from the HazCam using Theia structure-from-motion \cite{sweeney2023theia}, and then fuses the depth point clouds into a mesh \cite{beyer2018asp}. This method was used to reconstruct a 3D model of the Granite Lab environment included with the contributed dataset and used throughout this work.

\subsubsection{Computing Inconsistencies}
\label{sec:inconsistencies}

\begin{figure*}
    \centering
    \footnotesize
    \includegraphics[width=\textwidth]{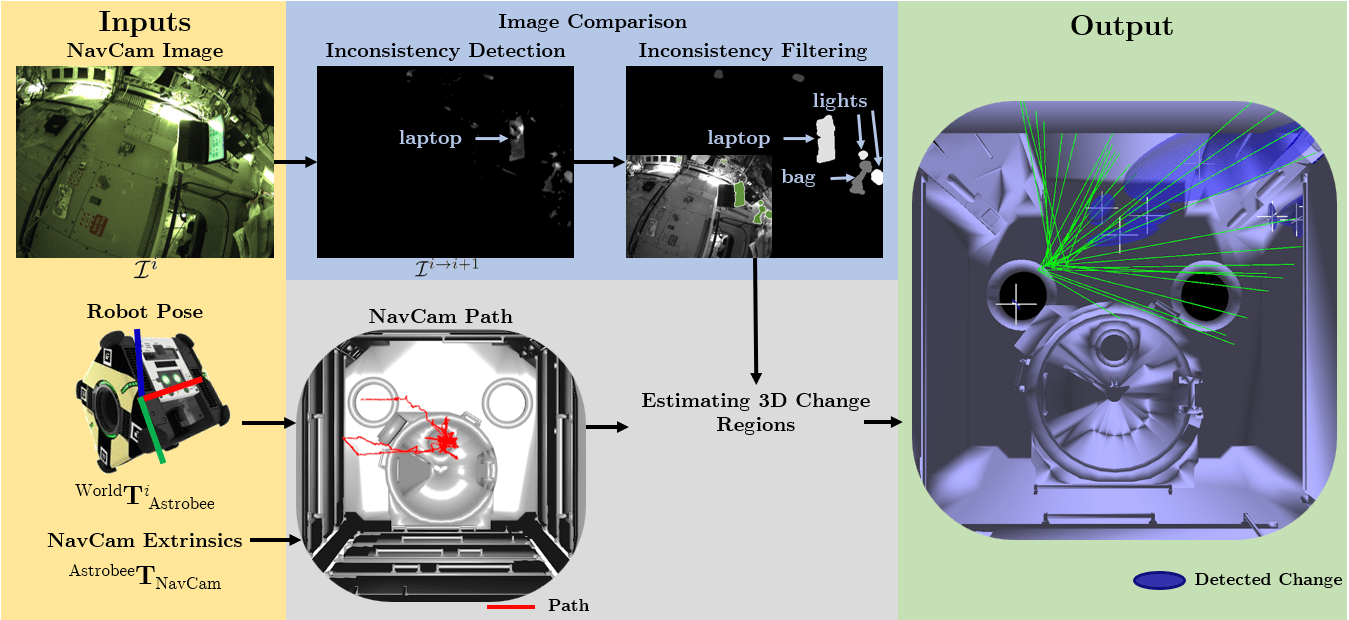}
    \caption{\justifying A laptop and lights are discovered in the ISS from a set of $n=7$ Astrobee NavCam images. (Left) FastCD inputs include a sequence of $n=7$ images, $\mathcal{I}^i$, and the pose at which each image was captured, $\mat{T}^i$. (Center Top) FastCD computes inconsistencies between images given the 3D model and filters the inconsistency image. (Center Bottom) Images are projected onto a 3D model given the path and pose of the camera to estimate 3D change regions. (Right) The 3D change regions are projected into the 3D model as change ellipses.}
    \label{fig: inconsistencies}
    \vspace{-2em}
\end{figure*}

Given the camera projection matrix, $\mat{P}_{3\times4}^i$, the projection of an arbitrary 3D point in homogeneous coordinates, $\mat{X}^i_{4 \times 1}$, onto the plane of image $i$ as a 2D pixel in homogeneous coordinates, $\mat{x}^i_{3 \times 1}$, is

\begin{equation}
\label{eq: point-to-pixel}
    \mat{x}^i = \mat{P}^i \mat{X}^i,
\end{equation}

\noindent where the projection matrix

\begin{equation}
    \mat{P}^i=\mat{K} \mat{R}^i\left[\mat{I}_{3\times3} \mid -\mat{t}^i\right]
\end{equation} 

\noindent is computed from the camera intrinsic calibration matrix $\mat{K}_{3 \times 3}$ and the rotation $\mat{R}_{3 \times 3}^i$ and translation $\mat{t}_{3 \times 1}^i$ components of $\left(\mat{T}^i\right)^{-1}$ that transform the World coordinates into NavCam coordinates. Inverting Eq. (\ref{eq: point-to-pixel}) gives

\begin{equation}
\label{eq: view-direction}
    \mat{r}^i = \left(\mat{R}^{i}\right)^{\intercal}\mat{K}^{-1}\mat{x}^i,
\end{equation}

\noindent the ray from the projection center of the camera through pixel $\mat{x}^i$ to the World. Each pixel of image $\mathcal{I}^i_{1280\times960}$ can then be back-projected onto the 3D model.

Inconsistencies are detected between a pair of images captured from poses $\mat{T}^i$ and $\mat{T}^{i+1}$ by creating a new intensity image, $\mathcal{I}^{i\rightarrow i+1}_{1280 \times 960}$, to store the content of $\mathcal{I}^i$ from pose $\mat{T}^{i+1}$ given the 3D model. The intersection, $\mat{X}^{i\rightarrow i+1}$, between a ray $\mat{r}^i$ from Eq. (\ref{eq: view-direction}) and the 3D model is computed and and projected onto $\mathcal{I}^{i\rightarrow i+1}$ as

\begin{equation}
    \label{eq: projection}
    \mat{x}^{i \rightarrow i+1} = \mat{P}^{i+1} \mat{X}^{i\rightarrow i+1},
\end{equation}

\noindent where $\mat{P}^i$ is the camera projection matrix corresponding to pose $\mat{T}^i$. Since image $\mathcal{I}^i$ and image $\mathcal{I}^{i \rightarrow i+1}$ are at the same perspective, $\mathcal{I}^{i \rightarrow i+1}$ can be used to segment $\mathcal{I}^{i+1}$. 

An accurate camera pose is unavailable for most robotic systems and the 3D model may be noisy. The pixel $\mat{x}^{i \rightarrow i+1}$ therefore has an associated uncertainty $\mat{\Sigma}_{3\times3}$ defined as

\begin{equation}
    \mat{\Sigma} := \mat{\Sigma}_{\mat{x}^{i\rightarrow i+1}\mat{x}^{i\rightarrow i+1}}.
\end{equation}

\noindent To account for this uncertainty, for each $\mat{x}^{i+1} \in \mathcal{I}^{i+1}$ compute the minimum Euclidean norm of the
intensity difference to each pixel $\mat{z}$ of $\mathcal{I}^{i\rightarrow i+1}$ in a region, $\mathcal{R}$, around $\mat{x}^{i+1}$ as

\begin{equation}
    d^{i\rightarrow i+1}(\mat{x}^{i+1}) = \min_{\mat{z} \in \mathcal{R}} \| \mathcal{I}^{i+1}(\mat{x}^{i+1}) - \mathcal{I}^{i\rightarrow i+1}(\mat{z}) \|.
\end{equation}

\noindent For a distance threshold $\tau^2=11.82$, the critical value of the $\chi_2^2$ distribution corresponding to the $3\sigma$ bound on the standard normal distribution, the region $\mathcal{R}$ is defined as

\begin{equation}
\small
    \mathcal{R}(\mat{z}) =
    \begin{cases}
      \left(\mat{x}^{i+1} - \mat{z}\right)^{\intercal} \mat{\Sigma}^{-1} \left(\mat{x}^{i+1}
      - \mat{z}\right) < \tau^2 & \forall \mat{z} \in \mathcal{I}^{i\rightarrow i+1}  \\
      0 & \text{otherwise}
    \end{cases}.
\end{equation}

\noindent If no change is detected between the 3D model and images $\mathcal{I}^{i}$ and $\mathcal{I}^{i+1}$, each $\mat{x}^i \in \mathcal{I}^i$ should re-project onto $\mathcal{I}^{i+1}$, images $\mathcal{I}^{i\rightarrow i+1}$ and $\mathcal{I}^{i+1}$ should be identical, and $\forall \mat{x}^{i+1} \in {\mathcal{I}^{i+1}}$, $d^{i\rightarrow i+1}(\mat{x}^{i+1}) \approx 0$. If there is a mismatch between the 3D model and images due to a change in the environment, pixels corresponding to the change will re-project onto the wrong place in $\mathcal{I}^{i+1}$ and the change will be detected from a large $d^{i\rightarrow i+1}(\mat{x}^{i+1})$ value. 

As originally highlighted in FastCD, this method has ambiguities when only two images are compared. A point $\mat{X}_c^i$ corresponding to a change in the 3D model generates two pixel locations on $\mathcal{I}^{i \rightarrow i+1}$. FastCD resolves this ambiguity using multiple pair-wise image comparisons \cite{palazzolo2018geometric}. Two pixels $\mat{x}^{i+1}$ and $\mat{x}^{i+2}$ belonging to the same change will project onto $\mathcal{I}^{i}$ at the same location. Therefore, pixels from different images which re-project onto the same region of $\mathcal{I}^i$ represent the real locations of change. Change regions are estimated in 2D by comparing each image with $m$ neighboring images. The default value of $m=4$ is used in this work unless otherwise noted. An example illustrating how inconsistencies are computed is included in Figure \ref{fig: inconsistencies}. After an inconsistency image with the 2D regions of change is computed, the inconsistency image is filtered through a sequence of image processing steps including erosion-dilation, contour filtering, and change region association.

\subsubsection{Regions of Change}

Once the regions containing changes are identified in 2D, the corresponding mean location of change, $\overline{\mat{X}}_c^i$, is estimated in 3D \cite{palazzolo2018geometric, forstner2016photogrammetric}. Every region identified as a change has mean pixel location $\overline{\mat{x}}^i$ and covariance $\mat{\Sigma}^i$. The corresponding 3D point $\overline{\mat{X}}_c^i$ is computed in World coordinates by triangulating the mean location of change in each image. The 3D locations of change are estimated by solving

\begin{equation} 
\label{eq: system}
\begin{matrix}
\mat{A} \overline{\mat{X}}_c=0,
\end{matrix}
\end{equation}

\noindent In Eq. (\ref{eq: system}),

\begin{equation}
\mat{A}_{3n\times4}=\begin{bmatrix} \mathrm{S}\left(\overline{\mat{x}}^{1}\right)\mat{P}^{1}, & \hdots, & \mathrm{S}\left(\overline{\mat{x}}^{n}\right)\mat{P}^{n} \end{bmatrix}^{\intercal},
\end{equation}

\noindent $n$ is the number of images, $\mat{P}^i$ is the projection matrix corresponding to pose $\mat{T}^i$, and $\mathrm{S}\left(\overline{\mat{x}}^i\right)$ is a skew-symmetric matrix generated from the elements of $\overline{\mat{x}}^i = \left[x^i, y^i, w^i\right]^{\intercal}$ such that

\begin{equation} 
\mathrm{S}(\overline{\mat{x}}^i)=\begin{bmatrix} 0 & -w^i & y^i\\ w^i & 0 & -x^i\\ -y^i & x^i & 0 \end{bmatrix}.
\end{equation}

\noindent Eq. (\ref{eq: system}) is solved using singular value decomposition~\cite{palazzolo2018geometric}. This procedure is repeated for the sigma points from every mean and covariance of the same region in every image to efficiently estimate 3D change regions without requiring a dense reconstruction of the environment. The mean of the points $\overline{\mat{X}}_c$ represent the regions in 3D where changes occurred. The covariances associated with $\overline{\mat{X}}_c$ are used to draw covariance ellipses to characterize the uncertainty of the estimate.

\subsection{Results and Limitations}
\label{sec:results}

FastCD was applied to data collected by Astrobee units in two environments. The ``Bsharp'' Astrobee unit surveyed the controlled Granite Lab environment (Section \ref{sec: granite-lab}) and the ``Bumble'' and ``Queen'' Astrobee units surveyed the uncontrolled ISS environment (Section \ref{sec: iss}). In this qualitative study, FastCD detects large objects appearing in multiple sequential image frames.

\vspace{-1em}
\subsubsection{Granite Lab Results}
\label{sec: granite-lab}

\noindent The ``Bsharp'' Astrobee unit surveyed the Granite Lab five times. The robot followed mostly translational paths, scanning each wall of the Granite Lab. The first four surveys (\texttt{20230419\_bsharp\_survey1}, \texttt{20230419\_bsharp\_survey2}, \texttt{20230419\_bsharp\_survey3}, and \texttt{20230419\_bsharp\_survey4}) are included in the released dataset. For each of these surveys, one or more confounding objects realistic to the ISS environment are introduced, including a cargo bag, a crate, a cable, and a static Astrobee unit, shown in Figure~\ref{fig: gt}. The fifth survey is not included in the released dataset. This survey was used to build the 3D model (\texttt{model.obj}) included with each Granite Lab survey, following the framework described in Section \ref{sec:baseline-map}. Object discovery in Granite Lab data is shown in Figure~\ref{fig: granitedata}. 

\columnbreak

\begin{figurehere}
    \centering
    \footnotesize\captionsetup{type=figure}\includegraphics[width=0.8\columnwidth]{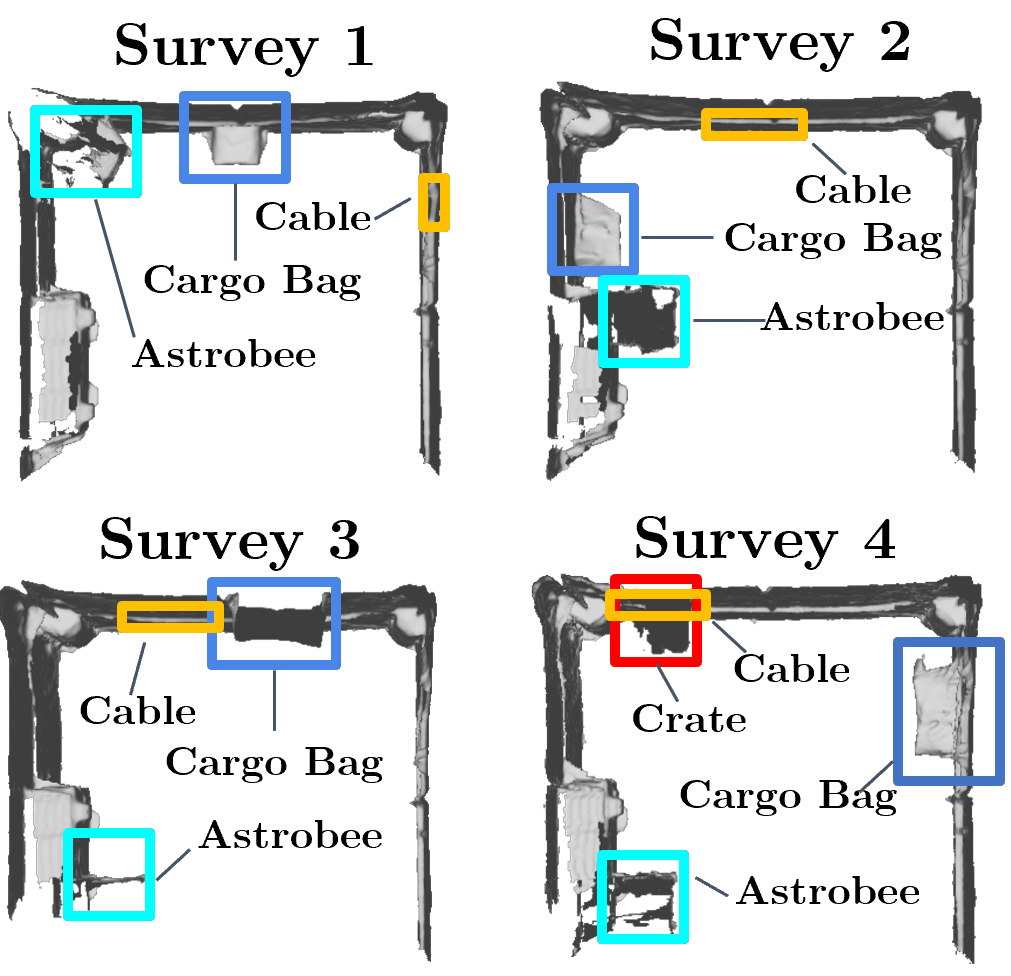}\caption{\justifying Objects are introduced at known locations to the Granite Lab environment for each of four surveys.}
    \label{fig: gt}
    \vspace{1em}
\end{figurehere}

\begin{figurehere}
\centering
\captionsetup{type=figure}
 \includegraphics[width=0.45\textwidth]{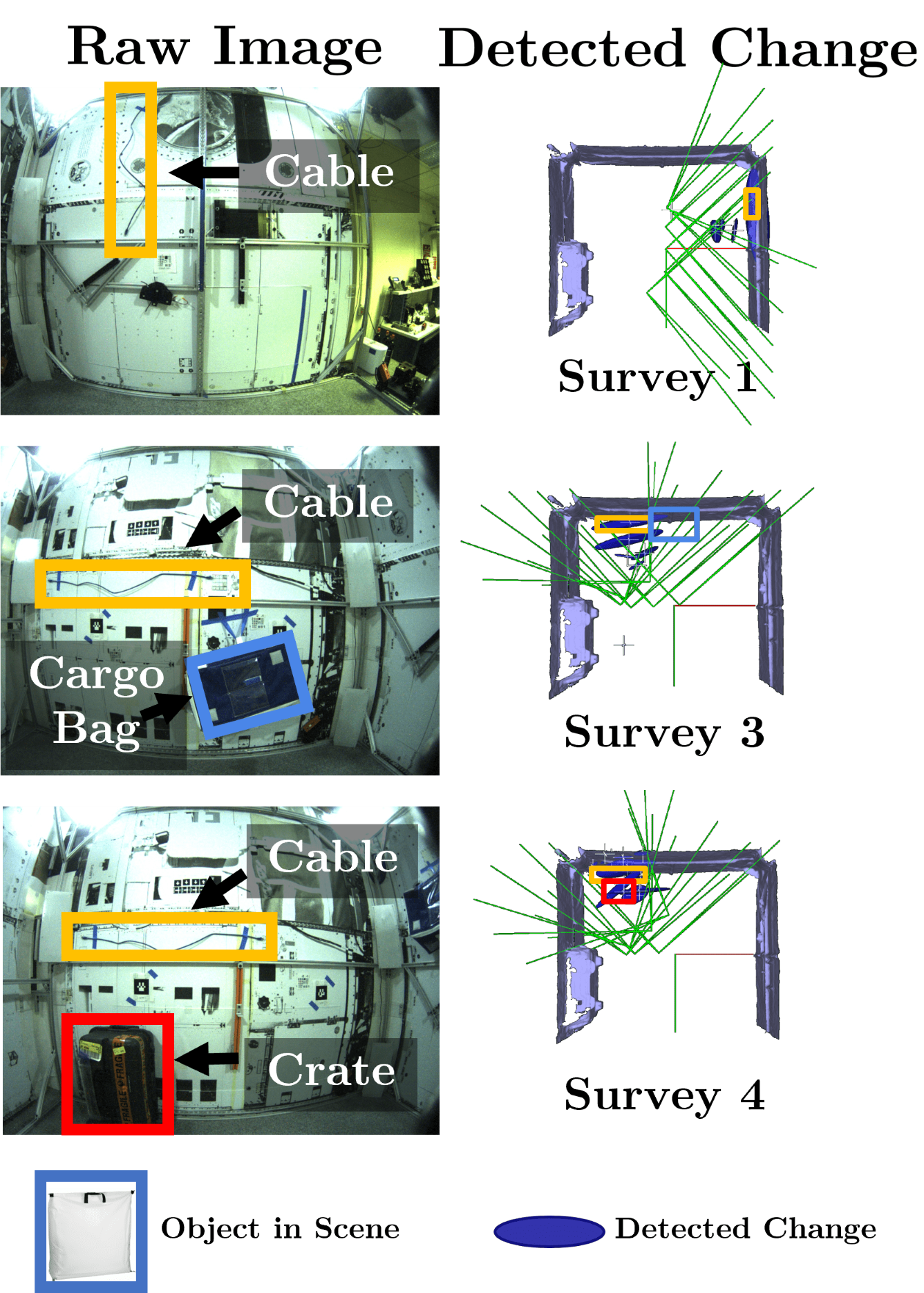}
\caption{\justifying FastCD detects several objects added to the Granite Lab environment. (Top Row) Cable detection from Survey 1. (Middle Row) Cable and cargo bag detection from Survey 3. (Bottom Row) Cable and crate detection from Survey 4.}
\label{fig: granitedata}
\vspace{1em}
\end{figurehere}

For each survey, the maximum number of between-image comparisons, $m$, was varied from 2 to 6, using the FastCD default $m = 4$ as a center point. Results were consistent with those found in \cite{palazzolo2018geometric}: with $m < 4$, the results were much noisier with increased false-positive detections. However, with each Granite Lab survey, the quality of the results did not increase with $m \geq 4$. False positives were occasionally found near the camera position, although less frequently, with $m \geq 4$. While the robot itself is technically a changing object between images in the scene, this result could be mitigated by filtering results within a certain proximity to the camera. After removing artifacts, these early FastCD results appear promising.

To analyze the accuracy of the results, the 3D coordinates of the known changes and of the detected changes are necessary ground truth data. The latter data are difficult to evaluate directly. Furthermore, the output model provides an overview of the area where changes are detected as a cluster of ellipses, and therefore a single item may be highlighted as several changes. However, for all surveys, the majority of ellipses are clustered around the location of the added object. 

Compute-constrained processors, including the 4-core, 2.5 [GHz] Snapdragon 805-based IFC6501s used by the Astrobee units, must meet real-time rates for autonomy software while running multiple processes concurrently \cite{smith2016astrobee}. Runtime must be verified to show, at the very least, standalone real-time performance of any autonomy algorithm. Computational timing data were collected for the first Granite Lab survey data on a computer with a 4-core, 3.20 [GHz] Intel i5-4570 CPU. Table \ref{tab:timing} reports runtimes for three FastCD computational processes based on the number of images, $n$, in the batch used to compute changes. All other FastCD parameters were kept at defaults. Data loading accounts for most of the runtime. Data are loaded as \texttt{.png} images and \texttt{.xml} poses and camera intrinsic parameters. For online change detection, these data could be retrieved from queues and made available to other processes such as localization. 

\begin{center}
\begin{tablehere}
\centering
\footnotesize
\captionsetup{type=table}
\caption{FastCD Runtime [s]}
\footnotesize
\begin{tabular}{|c|c|c|c|}
        \hline
        $n$ & Data Loading & Inconsistencies & 3D Change \\
        \hline
        2 & 0.844 & 0.311 & 0.054 \\ 
        3 & 0.800 & 0.606 & 0.076 \\ 
        4 & 0.843 & 1.156 & 0.134 \\
        5 & 0.835 & 1.891 & 0.161 \\
        6 & 0.842 & 2.280 & 0.266 \\
        \hline
        Per Image & -- & $\approx$ 0.281 & -- \\
        \hline
\end{tabular}
\newline
\label{tab:timing}
\end{tablehere}
\end{center}

The most computationally intensive component of FastCD is computing inconsistencies, with computation time scaling with $n$. For $n<7$, FastCD localizes scene changes in $<2.5\mathrm{s}$, fast enough to present actionable information about the environment to a mobile robot. However, for a change detection algorithm running on Astrobee, a $10\times$ slowdown is expected due to the computational load of other processes. Real-time performance of FastCD on Astrobee units is expected to require further optimization.

\subsubsection{ISS Activity Results}
\label{sec: iss}

The ``Bumble'' and ``Queen'' Astrobee units each performed three surveys of the ISS. The robots followed mostly rotational paths to scan the ISS Japanese Experiment Module (JEM) autonomously. All six of these surveys are included inh released dataset (\texttt{20220608\_bumble\_survey1}, \texttt{20220608\_bumble\_survey2}, \texttt{20220608\_bumble\_survey3}, \texttt{20220608\_queen\_survey1}, \texttt{20220608\_queen\_survey2}, and \texttt{20220608\_queen\_survey3}). Unlike in the Granite Lab, the environment was not modified for the purpose of this work: the ISS was surveyed as-is. The complexity of the ISS environment precludes reconstruction of a clean map from robot survey data, so this work uses the Astrobee simulation JEM model~\cite{nasa-description}. Samples of FastCD object discovery on the ISS data are shown in Figure~\ref{fig: max_comparisons}. 

The results in Figure~\ref{fig: max_comparisons}I  The ISS, like many human habitats, is cluttered. This makes it more difficult to detect changes at both local and full-scene scales. The FastCD algorithm uses image processing functions to remove insignificant regions estimated as change regions. Increasing the number of geometric projection operations by increasing $m$ also reduces noise. Future work using semantic labeling could further reduce noise prior to 3D location of change estimation. It is also important to note that the 3D Model used for the ISS Activities is a simulated model of the JEM and was not generated from image and pose data in the same way the Granite Lab 3D model was generated. Since the model was not reconstructed from real data, it introduces additional uncertainties due to the Sim2Real gap.

\begin{figurehere}
    \vspace{1em}
    \centering
    \footnotesize
    \captionsetup{type=figure}
    \includegraphics[width=\columnwidth]{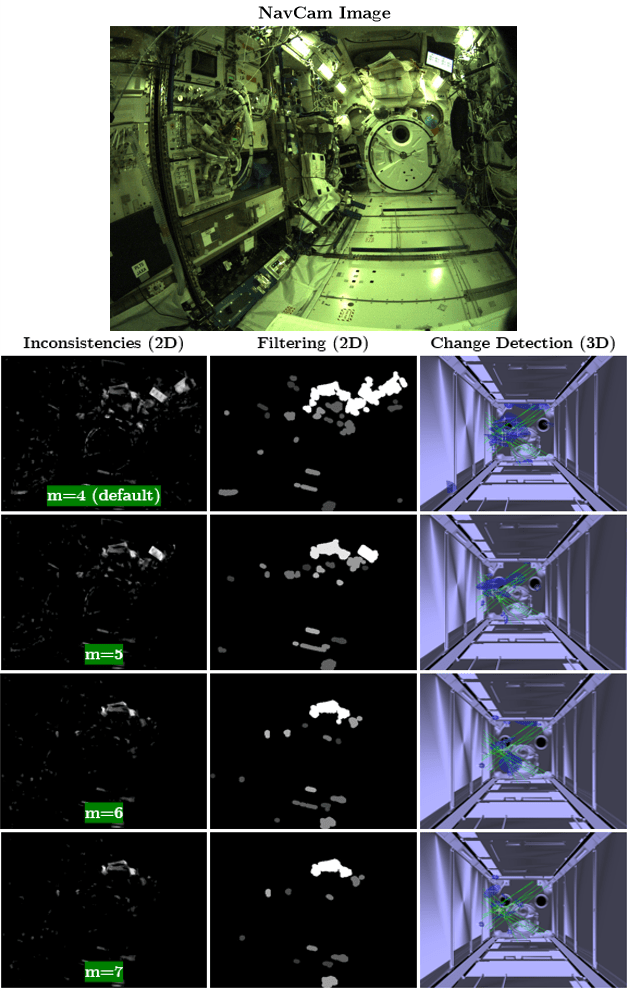}
    \caption{\justifying As the maximum number of between-image comparisons, $m$, increases, the number of detected changes decreases, removing spurious detections.}
    \label{fig: max_comparisons}
\end{figurehere}

\subsubsection{Limitations}

FastCD assumes a good estimate of the camera pose. One way this limitation can be addressed for Astrobee is to perform bundle adjustment against a known 3D map of the environment to obtain an accurate ground truth localization. For offline applications, this method of obtaining the ground truth pose may be acceptable. However, for online surveying and change detection, Astrobee would not have access to ground truth pose information and an inaccurate camera pose could lead to false positive or undetected changes.

Astrobee localization accuracy suffers most when the movement between consecutive images is low and when the number of features shared between consecutive images is low. This is usually due to rotation-only movement with little or no translation, introducing cheirality issues \cite{soussan2022astroloc}. As illustrated in Fig. \ref{fig: paths}, the Granite Lab surveys used T-shaped paths designed to increase translation in the survey. The ISS Activity surveys used rotate-in-place paths. These rotation-only surveys suffer the most from localization inaccuracy, increasing change detection errors. Other hardware- and environment-specific requirements include good lighting, prominent visual features, and a sufficient field of view for features to overlap between frames.

\subsection{Discussion and Future Work}
Motivated by the promise of assistive robotic technology in space environments, this work applied FastCD, a fast scene change detection algorithm, to object discovery for habitat maintenance. After discussing the specific considerations for scene change detection on a resource-constrained mobile robot in a space environment, FastCD was demonstrated on data collected with Astrobee units in the Granite Lab and aboard the ISS. The dataset used in this work is publicly released at \href{https://bit.ly/astrobee_fastcd_data}{https://bit.ly/astrobee\_fastcd\_data}.  

Future work could perform geometric scene change detection in real-time with multiple cameras, use images with semantic annotations to improve change localization, or feedback detected changes in real-time into an anomaly recovery architecture \cite{miller2022robust}. Future work could also improve on elements presented here. Currently, the limitations of FastCD for anomaly detection include sensitivity to object size, sensitivity to camera pose accuracy, and sensitivity to camera resolution. Robustness could also be further improved through detection pruning. 

Future work may also include runtime analysis on embedded processors. This work found computing inconsistencies is the most resource-consuming process and is related to $n$, the number of images in a batch. Computation time should also depend on $m$, the maximum number of comparisons used for detection in the FastCD algorithm. Investigating the impact of $m$ on timing will be important to keep computational expense within the resource budget of free-flyers such as Astrobee. Interesting extensions which may improve the accuracy of the FastCD algorithm include change detection labeling and semantics as well as robustness to illumination change.

Finally, this type of work unlocks numerous benefits for assistive robotics. Because FastCD estimates locations of change with uncertainty, when an environment map needs updating, only the local changes need to be identified and remapped. This feature is applicable to any robotic system, whether in space or terrestrial. Moreover, methods like FastCD allow Astrobee or other free-flyers to feasibly work alongside astronauts in space habitats, thus reducing astronaut workload. Furthermore, a great benefit of FastCD is its speed, which is especially useful for a resource-constrained mobile platform. As the world increasingly turns to automation to meet its needs, it is more important than ever to enable robots with the capability of responding to dynamic, human environments. 

\vspace{-1em} \subsection*{Acknowledgements}

\footnotesize 

The NASA Space Technology Graduate Research Opportunities 80NSSC21K1292 and 80NSSC18K1197 supported Holly Dinkel and Julia Di, respectively. Holly Dinkel was additionally supported by a P.E.O. Scholar Award and the Zonta International Amelia Earhart Fellowship. Julia Di was additionally supported by the Soffen Memorial Fund, Stanford JEDI Initiative, and Future Space Leaders Foundation. A portion of this research was carried out at the Jet Propulsion Laboratory, California Institute of Technology, under a contract with the National Aeronautics and Space Administration (80NM0018D0004). The CSIRO Data61 and the Winston Churchill Fellowship supported Paulo V.K. Borges. The NASA Game Changing Development (GCD) Program supported Jamie Santos, Marina Moreira, Oleg Alexandrov, Brian Coltin, and Trey Smith. The authors thank Ryan Soussan and the Astrobee Facilities team for supporting this work.

\renewcommand\bibsection{\subsection*{References}}
\bibliographystyle{iac}
\bibliography{natureabrv, references}

\begin{thebibliography}{10}
\providecommand{\url}[1]{#1}
\csname url@samestyle\endcsname
\providecommand{\newblock}{\relax}
\providecommand{\bibinfo}[2]{#2}
\providecommand{\BIBentrySTDinterwordspacing}{\spaceskip=0pt\relax}
\providecommand{\BIBentryALTinterwordstretchfactor}{4}
\providecommand{\BIBentryALTinterwordspacing}{\spaceskip=\fontdimen2\font plus
\BIBentryALTinterwordstretchfactor\fontdimen3\font minus
  \fontdimen4\font\relax}
\providecommand{\BIBforeignlanguage}[2]{{%
\expandafter\ifx\csname l@#1\endcsname\relax
\typeout{** WARNING: IEEEtran.bst: No hyphenation pattern has been}%
\typeout{** loaded for the language `#1'. Using the pattern for}%
\typeout{** the default language instead.}%
\else
\language=\csname l@#1\endcsname
\fi
#2}}
\providecommand{\BIBdecl}{\relax}
\BIBdecl

\bibitem{crusanDeepSpaceGateway2018}
J.~C. Crusan, R.~M. Smith, D.~A. Craig, J.~M. Caram, J.~Guidi, M.~Gates, J.~M.
  Krezel, and N.~B. Herrmann,
  ``\href{https://ieeexplore.ieee.org/document/8396541}{Deep Space Gateway
  Concept: Extending Human Presence into Cislunar Space},'' \emph{{IEEE}
  Aerosp. Conf. Proc. (AERO)}, vol. 2018-March, pp. 1--10, 2018.

\bibitem{lehnhardt2022gateway}
E.~Lehnhardt, S.~Fuller, J.~Quasny, D.~Connell, C.~Zaid, and K.~Halloran,
  ``\href{https://ntrs.nasa.gov/api/citations/20220013482/downloads/Gateway%20building%20block%20IAC-22%2CD3%2C1%2C1%2Cx69482_FINAL.pdf}{The
  Gateway as a Building Block for Space Exploration and Development},'' in
  \emph{{IAF} Int. Astr. Cong.}, vol.~73, 2022.

\bibitem{otero2002spheres}
A.~S. Otero, A.~Chen, D.~W. Miller, and M.~Hilstad,
  ``\href{https://ieeexplore.ieee.org/document/1036828}{SPHERES: Development of
  an ISS Laboratory for Formation Flight and Docking Research},'' in
  \emph{{IEEE} Aerosp. Conf. Proc. (AERO)}, vol.~1, 2002, pp. 59--73.

\bibitem{bualatAstrobeeNewTool2018}
M.~G. Bualat, T.~Smith, E.~E. Smith, T.~Fong, and D.~W. Wheeler,
  ``\href{https://ntrs.nasa.gov/api/citations/20180003326/downloads/20180003326.pdf}{Astrobee:
  A New Tool for ISS Operations},'' in \emph{AIAA Int. Conf. Space Ops.},
  vol.~15, May 2018, pp. 1--11.

\bibitem{smith2016astrobee}
T.~Smith, J.~Barlow, M.~Bualat, T.~Fong, C.~Provencher, H.~Sanchez, and
  E.~Smith,
  ``\href{https://ntrs.nasa.gov/api/citations/20160007769/downloads/20160007769.pdf}{Astrobee:
  A New Platform for Free-Flying Robotics on the International Space
  Station},'' in \emph{Int. Sympos. Artif. Intell. Robot. Autom. in Space
  (i-SAIRAS)}, 2016.

\bibitem{carlino2019astrobee}
R.~Carlino, J.~Barlow, J.~Benavides, M.~Bualat, A.~Katterhagen, Y.~Kim,
  R.~Garcia~Ruiz, T.~Smith, and A.~Mora~Vargas,
  ``\href{https://ntrs.nasa.gov/api/citations/20190032508/downloads/20190032508.pdf}{Astrobee
  Free Flyers: Integrated and Tested. Ready for Launch!}'' \emph{{IAF} Int.
  Astr. Cong.}, vol.~70, pp. 1--9, 2019.

\bibitem{bualat2021astrobee}
M.~G. Bualat, J.~S. Barlow, J.~V. Benavides, B.~Coltin, L.~Fl{\"u}ckiger,
  M.~Gouveia~Moreira, K.~M. Hamilton, A.~Katterhagen, R.~Soussan, T.~Smith, and
  T.~A. Team,
  ``\href{https://ntrs.nasa.gov/api/citations/20210014015/downloads/SpaceOps2021_ID1518_final.pdf}{Astrobee
  On-Orbit Commissioning},'' \emph{AIAA Int. Conf. Space Ops.}, vol.~16, pp.
  1--9, 2021.

\bibitem{astrobee}
NASA, ``{Astrobee Robot Software},'' [Online] Available:
  \href{https://github.com/nasa/astrobee}{https://github.com/nasa/astrobee},
  2023.

\bibitem{palazzolo2017change}
E.~Palazzolo and C.~Stachniss,
  ``\href{https://www.ipb.uni-bonn.de/wp-content/papercite-data/pdf/palazzolo2017irosws.pdf}{Change
  Detection in 3D Models Based on Camera Images},'' in \emph{IEEE/RSJ Int.
  Conf. on Intell. Robot. Sys. (IROS) Workshop on Planning, Perception and
  Navigation for Intelligent Vehicles}, 2017.

\bibitem{qin20163d}
R.~Qin, J.~Tian, and P.~Reinartz,
  ``\href{https://www.sciencedirect.com/science/article/pii/S0924271616304026}{3D
  Change Detection--Approaches and Applications},'' \emph{ISPRS J. Photogramm},
  vol. 122, pp. 41--56, 2016.

\bibitem{alcantarilla2018street}
P.~F. Alcantarilla, S.~Stent, G.~Ros, R.~Arroyo, and R.~Gherardi,
  ``\href{https://link.springer.com/article/10.1007/s10514-018-9734-5}{Street-View
  Change Detection with Deconvolutional Networks},'' \emph{Auton. Robot.},
  vol.~42, pp. 1301--1322, 2018.

\bibitem{celik2009unsupervised}
T.~Celik, ``\href{https://ieeexplore.ieee.org/document/5196726}{Unsupervised
  Change Detection in Satellite Images Using Principal Component Analysis and
  $K$-Means Clustering},'' \emph{{IEEE} Geosci. Remote Sens. Lett.}, vol.~6,
  no.~4, pp. 772--776, 2009.

\bibitem{radke2005image}
R.~J. Radke, S.~Andra, O.~Al-Kofahi, and B.~Roysam,
  ``\href{https://ieeexplore.ieee.org/document/1395984}{Image Change Detection
  Algorithms: A Systematic Survey},'' \emph{{IEEE} Trans. Image Process.},
  vol.~14, no.~3, pp. 294--307, 2005.

\bibitem{santos2023unsupervised}
J.~Santos, H.~Dinkel, J.~Di, P.~V. Borges, M.~Moreira, B.~Coltin, and T.~Smith,
  ``\href{https://arc.aiaa.org/doi/10.2514/6.2024-1960}{Unsupervised Change
  Detection for Space Habitats Using 3D Point Clouds},'' in \emph{{AIAA}
  SciTech F.}, 2024.

\bibitem{li2009novelGMM}
Y.~Li and L.~Li, ``\href{https://ieeexplore.ieee.org/document/5365128}{A Novel
  Split and Merge EM Algorithm for Gaussian Mixture Model},'' in \emph{Int.
  Conf. Nat. Comput.}, vol.~6.\hskip 1em plus 0.5em minus 0.4em\relax IEEE,
  2009, pp. 479--483.

\bibitem{rubner2000earth}
Y.~Rubner, C.~Tomasi, and L.~J. Guibas,
  ``\href{https://link.springer.com/article/10.1023/A:1026543900054}{The Earth
  Mover's Distance as a Metric for Image Retrieval},'' \emph{Int. J. Comput.
  Vis.}, vol.~40, no.~2, p.~99, 2000.

\bibitem{kanji2019localization}
T.~Kanji,
  ``\href{https://ieeexplore.ieee.org/document/8793482}{Detection-by-Localization:
  Maintenance-Free Change Object Detector},'' in \emph{{IEEE} Int. Conf. Robot.
  Autom. (ICRA)}, 2019, pp. 4348--4355.

\bibitem{golparvarfard2011monitoring}
M.~Golparvar-Fard, F.~Peña-Mora, and S.~Savarese,
  ``\href{https://ieeexplore.ieee.org/document/6130250}{Monitoring Changes of
  3D Building Elements From Unordered Photo Collections},'' in \emph{{IEEE}
  Int. Conf. Comput. Vis. (ICCV)}, 2011, pp. 249--256.

\bibitem{taneja2011image}
A.~Taneja, L.~Ballan, and M.~Pollefeys,
  ``\href{https://ieeexplore.ieee.org/document/6126515}{Image Based Detection
  of Geometric Changes in Urban Environments},'' in \emph{{IEEE} Int. Conf.
  Comput. Vis. (ICCV)}.\hskip 1em plus 0.5em minus 0.4em\relax IEEE, 2011, pp.
  2336--2343.

\bibitem{taneja2013city}
------, ``\href{https://ieeexplore.ieee.org/document/6618866}{City-Scale Change
  Detection in Cadastral 3D Models Using Images},'' in \emph{{IEEE/CVF} Int.
  Conf. Comput. Vis. Pattern Recognit. (CVPR)}, 2013, pp. 113--120.

\bibitem{bouguet2001pyramidal}
J.-Y. Bouguet \emph{et~al.},
  ``\href{http://robots.stanford.edu/cs223b04/algo_tracking.pdf}{Pyramidal
  Implementation of the Affine Lucas-Kanade Feature Tracker: Description of the
  Algorithm},'' \emph{Intel Corporation}, vol.~5, no. 1-10, p.~4, 2001.

\bibitem{dellaert2012factor}
F.~Dellaert,
  ``\href{https://repository.gatech.edu/entities/publication/0c2ac17c-1df4-48fe-8532-8f746868934a}{Factor
  Graphs and GTSAM: A Hands-On Introduction},'' \emph{Georgia Institute of
  Technology, Tech. Rep}, vol.~2, p.~4, 2012.

\bibitem{carlone2014eliminating}
L.~Carlone, Z.~Kira, C.~Beall, V.~Indelman, and F.~Dellaert,
  ``\href{https://ieeexplore.ieee.org/document/6907483/}{Eliminating
  Conditionally Independent Sets in Factor Graphs: A Unifying Perspective Based
  on Smart Factors},'' in \emph{{IEEE} Int. Conf. Robot. Autom. (ICRA)}, 2014,
  pp. 4290--4297.

\bibitem{soussan2022astroloc}
R.~Soussan, V.~Kumar, B.~Coltin, and T.~Smith,
  ``\href{https://ieeexplore.ieee.org/document/9811919}{AstroLoc: An Efficient
  and Robust Localizer for a Free-flying Robot},'' in \emph{{IEEE} Int. Conf.
  Robot. Autom. (ICRA)}, 2022, pp. 4106--4112.

\bibitem{kalibr}
E.~A.~S. Laboratory, ``{Kalibr},'' [Online] Available:
  \href{https://github.com/ethz-asl/kalibr}{https://github.com/ethz-asl/kalibr},
  2023.

\bibitem{furgale2012kalibr}
P.~Furgale, T.~D. Barfoot, and G.~Sibley,
  ``\href{https://ieeexplore.ieee.org/document/6225005}{Continuous-Time Batch
  Estimation Using Temporal Basis Functions},'' in \emph{{IEEE} Int. Conf.
  Robot. Autom. (ICRA)}, 2012, pp. 2088--2095.

\bibitem{furgale2013kalibr}
P.~Furgale, J.~Rehder, and R.~Siegwart,
  ``\href{https://ieeexplore.ieee.org/document/6696514}{Unified Temporal and
  Spatial Calibration for Multi-Sensor Systems},'' in \emph{{IEEE/RSJ} Int.
  Conf. Intell. Robot. Sys. (IROS)}, 2013, pp. 1280--1286.

\bibitem{maye2013self}
J.~Maye, P.~Furgale, and R.~Siegwart,
  ``\href{https://ieeexplore.ieee.org/document/6629513}{Self-Supervised
  Calibration for Robotic Systems},'' in \emph{IEEE Intell. Veh. Symp. (IV)},
  2013, pp. 473--480.

\bibitem{smith2021isaac}
T.~Smith, M.~Bualat, A.~Akanni, O.~Alexandrov, L.~Barron, J.~Benton, B.~Coltin,
  T.~Fong, J.~Garcia, K.~Hamilton, L.~Hill, M.~Moreira, R.~Morris, N.~Ortega,
  J.~Pea, J.~Rogers, M.~Savchenko, K.~Sharif, and R.~Soussan,
  ``\href{https://longhorizon.org/trey/papers/smith21_isaac.pdf}{ISAAC: An
  Integrated System for Autonomous and Adaptive Caretaking},'' in \emph{ISS
  R\&D Conference}, 2021.

\bibitem{nasa-isaac}
NASA, ``{ISAAC (Integrated System for Autonomous and Adaptive Caretaking)},''
  [Online] Available:
  \href{https://github.com/nasa/isaac}{https://github.com/nasa/isaac}, 2023.

\bibitem{sweeney2023theia}
C.~Sweeney, ``{Theia Multiview Geometry Library: Tutorial \& Reference},''
  [Online] Available: \href{http://theia-sfm.org}{http://theia-sfm.org}, 2023.

\bibitem{beyer2018asp}
R.~A. Beyer, O.~Alexandrov, and S.~McMichael,
  ``\href{https://agupubs.onlinelibrary.wiley.com/doi/full/10.1029/2018EA000409}{The
  Ames Stereo Pipeline: NASA’s Open Source Software for Deriving and
  Processing Terrain Data},'' \emph{Earth Space Sci.}, vol.~5, 2018.

\bibitem{palazzolo2018geometric}
E.~Palazzolo and C.~Stachniss,
  ``\href{https://ieeexplore.ieee.org/document/8461019}{Fast Image-Based
  Geometric Change Detection Given a 3D Model},'' in \emph{{IEEE} Int. Conf.
  Robot. Autom. (ICRA)}, 2018, pp. 6308--6315.

\bibitem{forstner2016photogrammetric}
W.~F{\"o}rstner and B.~P. Wrobel,
  \emph{\href{https://link.springer.com/book/10.1007/978-3-319-11550-4}{Photogrammetric
  Computer Vision}}.\hskip 1em plus 0.5em minus 0.4em\relax Springer, 2016.

\bibitem{nasa-description}
NASA, ``{Astrobee Media},'' [Online] Available:
  \href{https://github.com/nasa/astrobee\_media.git}{https://github.com/nasa/astrobee\_media.git},
  2023.

\bibitem{miller2022robust}
I.~D. Miller, R.~Soussan, B.~Coltin, T.~Smith, and V.~Kumar,
  ``\href{https://ieeexplore.ieee.org/document/9811862}{Robust Semantic Mapping
  and Localization on a Free-Flying Robot in Microgravity},'' in \emph{{IEEE}
  Int. Conf. Robot. Autom. (ICRA)}, 2022, pp. 4121--4127.

\end{thebibliography}
\newpage
\normalsize
\subsection*{Authors}

\parpic{\includegraphics[width=1in,clip,keepaspectratio]{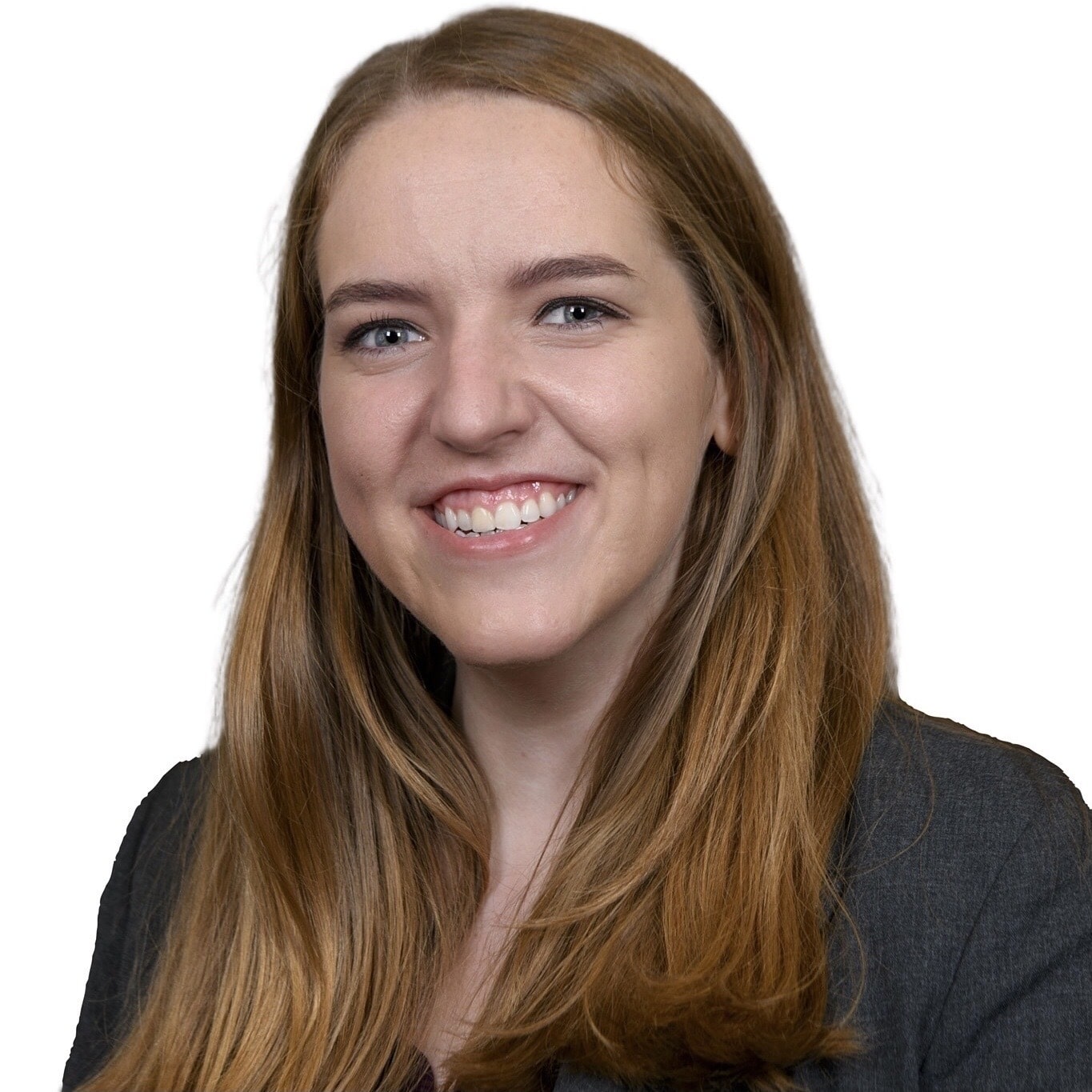}}
\noindent \textbf{Holly Dinkel} \href{https://www.linkedin.com/in/hollydinkel}{\faLinkedinSquare} \href{https://hollydinkel.github.io}{\faHome} is pursuing a Ph.D. in aerospace engineering at the University of Illinois Urbana-Champaign where she researches robotic caretaking as a NASA Space Technology Graduate Research Fellow with the NASA Ames Research Center Intelligent Robotics Group and the NASA Johnson Space Center Dexterous Robotics Laboratory.

\parpic{\includegraphics[width=1in,clip,keepaspectratio]{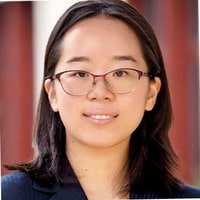}}
\noindent \textbf{Julia Di} \href{https://www.linkedin.com/in/JuliaDi}{\faLinkedinSquare} \href{https://web.stanford.edu/~juliadi/}{\faHome} is pursuing a Ph.D. in mechanical engineering at Stanford University where she researches tactile sensing and perception. She was a NASA Space Technology Graduate Research Fellow with the NASA Ames Research Center Intelligent Robotics Group and NASA Jet Propulsion Laboratory Mobility and Robotic Systems Section.

\parpic{\includegraphics[width=1in,clip,keepaspectratio]{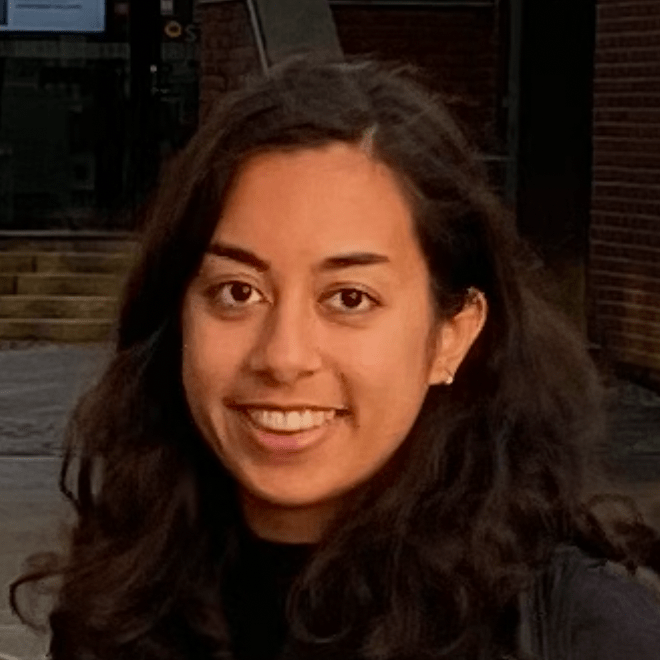}}
\noindent \textbf{Jamie Santos} \href{https://www.linkedin.com/in/jamiecsantos}{\faLinkedinSquare} completed a M.S. in Complex and Adaptive Systems at Chalmers University in Gothenburg, Sweden. She researches change detection to enable NASA’s free-flying Astrobee robots to detect anomalies on the International Space Station with the NASA Ames Research Center Intelligent Robotics Group.

\parpic{\includegraphics[width=1in,clip,keepaspectratio]{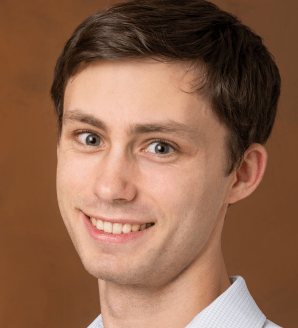}}
\noindent \textbf{Keenan Albee} \href{https://www.linkedin.com/in/kalbee}{\faLinkedinSquare} is a Robotics Technologist in the Maritime and Multi-Agent Autonomy group at the Jet Propulsion Laboratory, California Institute of Technology. He received a Ph.D. in Aeronautics and Astronautics from the Massachusetts Institute of Technology. He is interested in motion planning under uncertainty, information-aware planning, and real-time model learning.

\parpic{\includegraphics[width=1in,clip,keepaspectratio]{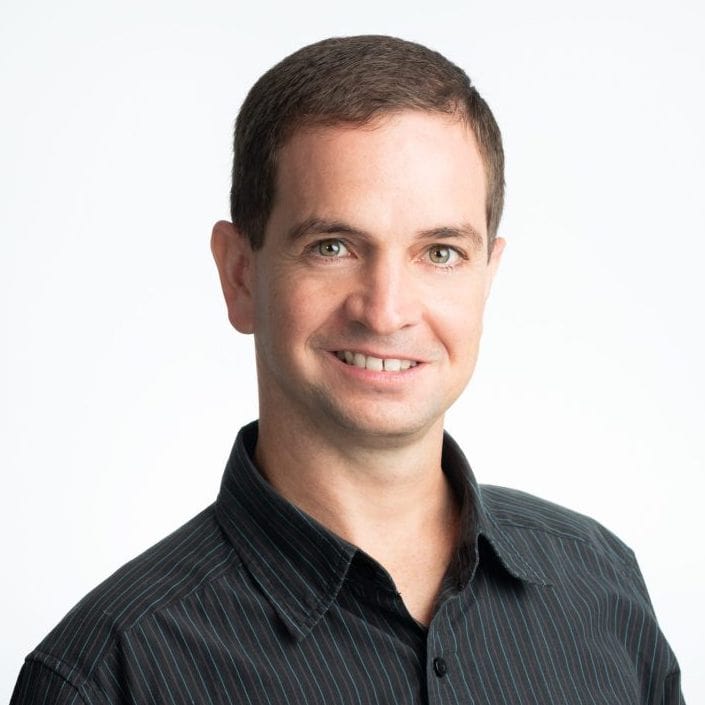}}
\noindent \textbf{Paulo Borges} \href{https://au.linkedin.com/in/paulovinicius}{\faLinkedinSquare} \href{https://paulovinicius.com/index.html}{\faHome} is a Principal Research Scientist in the Robotics and Autonomous Systems Group at CSIRO in Brisbane, Australia. He completed a Ph.D. in Electronic Engineering from Queen Mary University of London, London, United Kingdom. His research interests include robotic automation for the manufacturing, energy, and agriculture industries, bridging industry, innovation, and research. 

\columnbreak

\parpic{\includegraphics[width=1in,clip,keepaspectratio]{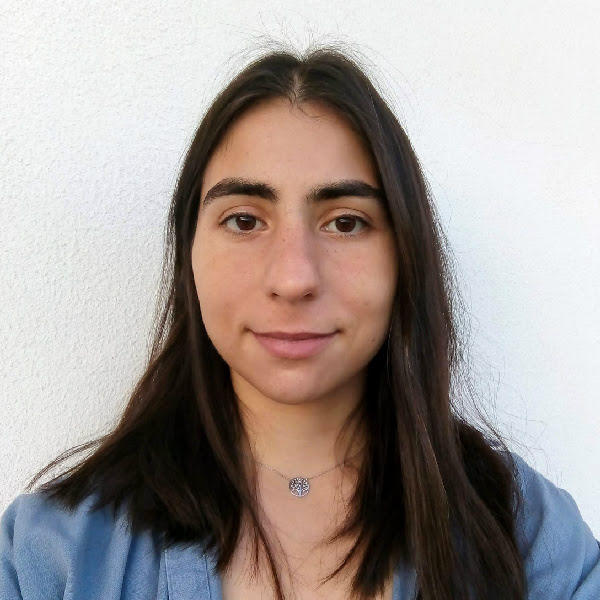}}
\noindent \textbf{Marina Moreira} \href{https://www.linkedin.com/in/marinagmoreira}{\faLinkedinSquare} is a research engineer in the NASA Ames Research Center Intelligent Robotics Group. She completed a M.S. in aerospace engineering from Instituto Superior Técnico, Lisbon, Portugal. She develops and maintains open-source software for the ISAAC project and is interested in problems related to robotic systems and control.

\parpic{\includegraphics[width=1in,clip,keepaspectratio]{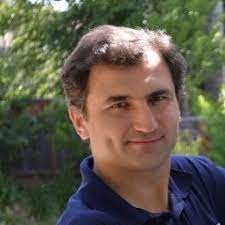}}
\noindent \textbf{Oleg Alexandrov} \href{https://linkedin.com/in/olegalexandrov}{\faLinkedinSquare} is a Research Scientist in the NASA Ames Research Center Intelligent Robotics Group. He received a Ph.D. in applied mathematics from the University of Minnesota. His research focuses on mapping using satellite and robot images.

\parpic{\includegraphics[width=1in,clip,keepaspectratio]{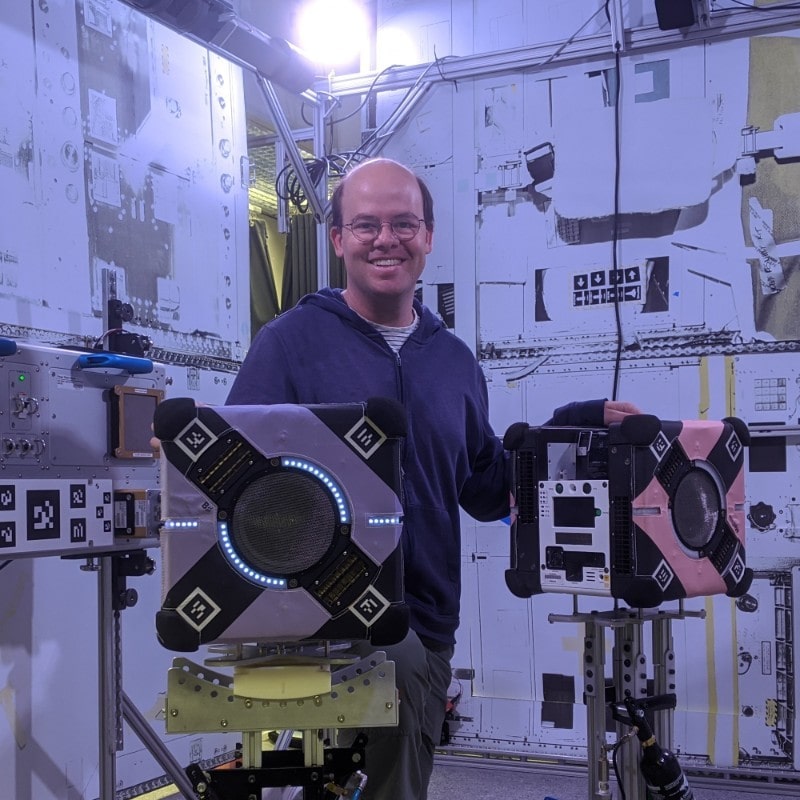}}
\noindent \textbf{Brian Coltin} \href{https://www.linkedin.com/in/bcoltin}{\faLinkedinSquare} \href{http://brian.coltin.org}{\faHome} is a Computer Scientist in the NASA Ames Research Center Intelligent Robotics Group where he works on the Astrobee robot, the VIPER rover, and flood mapping. He earned a Ph.D. in Robotics from Carnegie Mellon University. His research interests include planning, scheduling, multi-robot coordination, localization, and computer vision.

\parpic{\includegraphics[width=1in,clip,keepaspectratio]{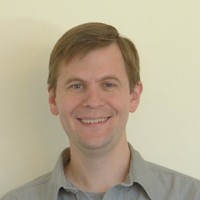}}
\noindent \textbf{Trey Smith} \href{https://www.linkedin.com/in/trey-smith-robotics}{\faLinkedinSquare} \href{https://longhorizon.org/trey/}{\faHome} is a Computer Scientist in the NASA Ames Research Center Intelligent Robotics Group where he works on the Astrobee robot and the VIPER rover. He earned a Ph.D. in Robotics from Carnegie Mellon University. His research interests include planning, scheduling, multi-robot coordination, mapping, and computer vision.

\end{multicols}

\end{document}